\documentclass[algorithms,article,accept,pdftex,moreauthors]{Definitions/mdpi} 

\usepackage{algorithmic}
\usepackage{algorithm}
\usepackage{listings}
\floatstyle{ruled}
\newfloat{listing}{tb}{lst}{}
\floatname{listing}{Listing}
\usepackage{makecell}

\firstpage{1} 
\pubvolume{1}
\issuenum{1}
\articlenumber{0}
\pubyear{2025}
\copyrightyear{2025}
\externaleditor{Frank Werner} 
\datereceived{27 August 2025} 
\daterevised{11 November 2025} 
\dateaccepted{22 November 2025  } 
\datepublished{ } 
\hreflink{https://doi.org/} 

\Title{Mxplainer: Explain and Learn Insights by Imitating\linebreak Mahjong Agents}

\TitleCitation{Mxplainer: Explain and Learn Insights by Imitating Mahjong Agents}

\Author{Lingfeng Li \orcidA{}, Yunlong Lu, Yongyi Wang, Qifan Zheng and Wenxin Li *}

\AuthorNames{Lingfeng Li, Yunlong Lu, Yongyi Wang, Qifan Zheng and Wenxin Li}

\isAPAStyle{%
       \AuthorCitation{Lastname, F., Lastname, F., \& Lastname, F.}
         }{%
        \isChicagoStyle{%
        \AuthorCitation{Lastname, Firstname, Firstname Lastname, and Firstname Lastname.}
        }{
        \AuthorCitation{Li, L.; Lu, Y.; Wang, Y.; Zheng, Q.; Li, W.}
        }
}

\address[1]{Peking University, Beijing 100871, China;
 {lingfengli@stu.pku.edu.cn (L.L.); luyunlong@pku.edu.cn (Y.L.); wangyongyi@pku.edu.cn (Y.W.); zqf1998@stu.pku.edu.cn (Q.Z.)} 
\\
}

\corres{\hangafter=1 \hangindent=1.05em \hspace{-0.82em}Correspondence: lwx@pku.edu.cn}

\abstract{People need to internalize the skills of AI agents to improve their own capabilities. Our paper focuses on Mahjong, a multiplayer game involving imperfect information and requiring effective long-term decision-making amidst randomness and hidden information. Through the efforts of AI researchers, several impressive Mahjong AI agents have already achieved performance levels comparable to those of professional human players; however, these agents are often treated as black boxes from which few insights can be gleaned.
This paper introduces Mxplainer, a parameterized search algorithm that can be converted into an equivalent neural network to learn the parameters of black-box agents. 
Experiments on both human and AI agents demonstrate that Mxplainer achieves a top-three action prediction accuracy of over 92\% and 90\%, respectively, while providing faithful and interpretable approximations that outperform decision-tree methods (34.8\% top-three accuracy). This enables Mxplainer to deliver both strategy-level insights into agent characteristics and actionable, step-by-step explanations for individual decisions.}


\begin{document}



\section{Introduction}

Games play a pivotal role in the field of AI, offering unique challenges to the research community and serving as fertile ground for the development of novel AI algorithms. Board games, such as Go~\cite{Silver2017} and chess~\cite{silver2017mastering}, provide ideal settings for perfect-information scenarios, where all agents are fully aware of the environment states. Card games, like heads-up no-limit hold’em (HUNL)~\cite{doi:10.1126/science.aao1733, doi:10.1126/science.aay2400}, Doudizhu~\cite{yang2024perfectdou, zha2021douzero}, and~Mahjong~\cite{li2020suphx}, present different dynamics with their imperfect-information nature, where agents must infer and cope with hidden information from states. Video games, such as Starcraft~\cite{Vinyals2019}, Minecraft~\cite{fan2022minedojo}, and~Honor of Kings~\cite{wei2022honor}, push AI algorithms to process and extract crucial features from a multitude of signals amidst~noise. 

Conversely, the~advancement of AI algorithms also incites new enthusiasm in games. The~work of AlphaGo~\cite{Silver2016} in 2016 has had a long-lasting impact on the Go community. It revolutionized the play style of Go, a~game with millennia of history, and~changed the perspectives of world champions on this game~\cite{aivictory}. Teaching tools based on AlphaGo~\cite{aiteach} have become invaluable resources for newcomers while empowering professional players to set new records~\cite{record2024}.

Mahjong, a~worldwide popular game with unique characteristics, has gained traction in the AI research community as a new testbed. It brings its own flavor as a multiplayer imperfect-information environment. First, there is no rank of individual tiles in Mahjong; all tiles are equal in their role within the game. A~game of Mahjong is not won by beating other players in card ranks but by being the first to reach a winning pattern. Therefore, state evaluation is more challenging for Mahjong players, as~they need to assess the similarity and distance between game states and the closest game goals. Second, the~game objective in Mahjong is to become the first to complete one of many possible winning patterns. The~optimal goal can frequently change when players draw new tiles during gameplay. In~fact, Mahjong's core difficulty lies in selecting the most effective goal among numerous possibilities. Players must evaluate their goals and make decisions upon drawing each tile and reacting to other players' discarded tiles. This decision-making process often involves situations where multiple goals are of similar distance, requiring trade-offs that distinguish play styles and reveal a player's level of~expertise.

\begin{figure}[H]
\includegraphics[width=0.5\columnwidth]{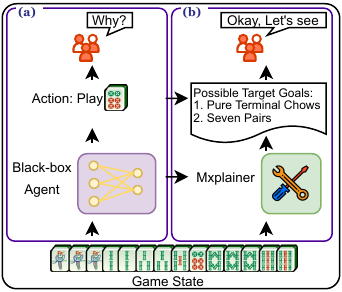} 
\caption{Comparison between current situation of black-box Mahjong agents and Mahjong agents with Mxplainer. (\textbf{a}). Raw action output without explanations. (\textbf{b}). Mxplainer explains possible reasons behind~actions. }
\label{figx}
\end{figure}

Several strong agents have already been developed for different variants of Mahjong rules~\cite{li2020suphx, 10.1109/CoG51982.2022.9893584, zhao2022building}. However, as~illustrated in Figure~\ref{figx}, without~the use of explainable AI methods~\cite{arrieta2019explainable, e23010018}, people can only observe the agents' actions without understanding how the game states are evaluated or which game goals are preferred that lead to those~actions.

In Explainable AI (XAI)~\cite{arrieta2019explainable}, black-box models typically refer to neural networks that lack inherent transparency and thus rely on post hoc explanation tools for interpretability~\cite{e23010018}. Existing XAI techniques, such as LIME~\cite{DBLP:journals/corr/RibeiroSG16} and LMUT~\cite{10.1007/978-3-030-10928-8_25}, struggle to provide both high-accuracy policy approximation and human-intelligible explanations in domains with immense state spaces and non-Markovian characteristics, a~gap that becomes critical for analyzing strategic agents in games like~Mahjong.

In this paper, we present Mxplainer (\textbf{{M}
}ahjong E\textbf{{xplainer}
}), a~parameterized classical agent framework designed to serve as an analytical tool for explaining the decision-making process of black-box Mahjong agents, as~shown in Figure~\ref{figx}b. Specifically, we have developed a parameterized framework that forms the basis for a family of search-based Mahjong agents. This framework is then translated into an equivalent neural network model, which can be trained using gradient descent to mimic any black-box Mahjong agent. Finally, the~learned parameters are used to populate the parameterized search-based agents. We consider these classical agents to be inherently explainable because each calculation and decision step within them is comprehensible to human experts. This enables detailed interpretation and analysis of the decision-making processes and characteristics of the original black-box~agents.

Through a series of experiments on game data from both AI agents and human players, we demonstrate that the learned parameters effectively reflect the decision processes of agents, including their preferred game goals and tiles to play. Our research also shows that by delving into the framework components, we can interpret the decision-making process behind the actions of black-box~agents.

This paper pioneers research on analyzing Mahjong agents by presenting Mxplainer, a~framework to explain black-box decision-making agents using search-based algorithms. Mxplainer allows AI researchers to profile and compare both AI agents and human players effectively. Additionally, we propose a method to convert any parameterized classical agent into a neural agent for automatic parameter tuning. Beyond~traditional approaches like decision trees, our work explores the conversion from neural agents to classical agents. 


The rest of this paper is organized as follows: We first review the related literature. Next, we introduce the rules of Mahjong and the specific variant used in our study. Then, we provide a detailed explanation of the components of our approach. Following this, we present a series of experiments demonstrating the effectiveness of our method in approximating and capturing the characteristics of different Mahjong agents. Finally, we discuss the implications of our work and conclude with future research~directions.

\section{Related~Works}

Explainable AI (XAI)~\cite{arrieta2019explainable} is a research domain dedicated to developing techniques for interpreting AI models for humans. This field encompasses several categories: classification models, generative AI, and~decision-making agents. A~specific subfield within this domain is Explainable Reinforcement Learning (XRL) \cite{10.1145/3616864}, which focuses on explaining the behavior of decision-making agents. Explanations in XRL can be classified into two main categories: global and~local.

Global explanations provide a high-level perspective on the characteristics and overall strategies of black-box agents, answering questions such as how an agent's strategy differs from others. Local explanations, on~the other hand, focus on the detailed decision-making processes of agents, elucidating why an agent selects action A over B under \mbox{specific~scenarios}. 

XRL methods can be classified into intrinsic and post hoc methods. Intrinsic methods directly generate explanations from the original black-box models, while post hoc methods rely on additional models to explain existing ones. Imitation learning (IL) \cite{10.1145/1015330.1015430, SCHAAL1999233} is a family of post hoc techniques that approximate a target policy. LMUT~\cite{10.1007/978-3-030-10928-8_25} constructs a U-tree with linear models as leaf nodes and approximates target policies through a node-splitting algorithm and gradient descent. Q-BSP~\cite{9207648} uses a batched Q-value to partition nodes and generate trees efficiently. EFS~\cite{Zhang2020} employs an ensemble of linear models with non-linear features generated by genetic programming to approximate target policies. These methods have been tested and excelled in environments such as CartPole and MountainCar, and~others have excelled in the Gym~\cite{brockman2016openai}. However, they rely on properly sampled state--action pairs to generate policies and may not be robust to out-of-distribution (OOD) states, which is particularly crucial for Mahjong with its high-dimensional state space and imperfect~information.

PIRL~\cite{pmlr-v80-verma18a} distinguishes itself among IL methods by introducing parameterized policy templates using its proposed policy programming language. It approximates the target $\pi$ through fitting parameters with Bayesian Optimization in Markov games, achieving high performance in the TORCS car racing game. Compared to TORCS, Mahjong has far more complex state features and requires encoding action histories within states to obtain the Markov property. Additionally, Mahjong agents must make multi-step decisions from game goal selection to tile picking. Similarly to PIRL, we define a parameterized search-based framework and optimize parameters using batched gradient descent to address these~challenges.

\section{Mahjong: the~Game}

Mahjong is a four-player imperfect information tile-based tabletop game. The~complexity of imperfect-information games can be measured by information sets, which are game states that players cannot distinguish from their own observations. The~average size of information sets in Mahjong is around $10^{48}$, making it a much more complex game to solve than Heads-Up Texas Hold'em~\cite{a16050235}, for which its average  information set size is around $10^3$. This immense complexity arises from the vast number of possible tile distributions, hidden player hands, and~the long sequence of interdependent decisions involving drawing, discarding, and~claiming tiles over multiple rounds. To~facilitate the readability of this paper, we highlight terminologies used in Mahjong with \textbf{{bold texts}
}, and~we distinguish scoring patterns (\textbf{fans}) by \textit{{italicized} 
 texts}.

\begin{figure}[H]
\includegraphics[width=0.65\columnwidth]{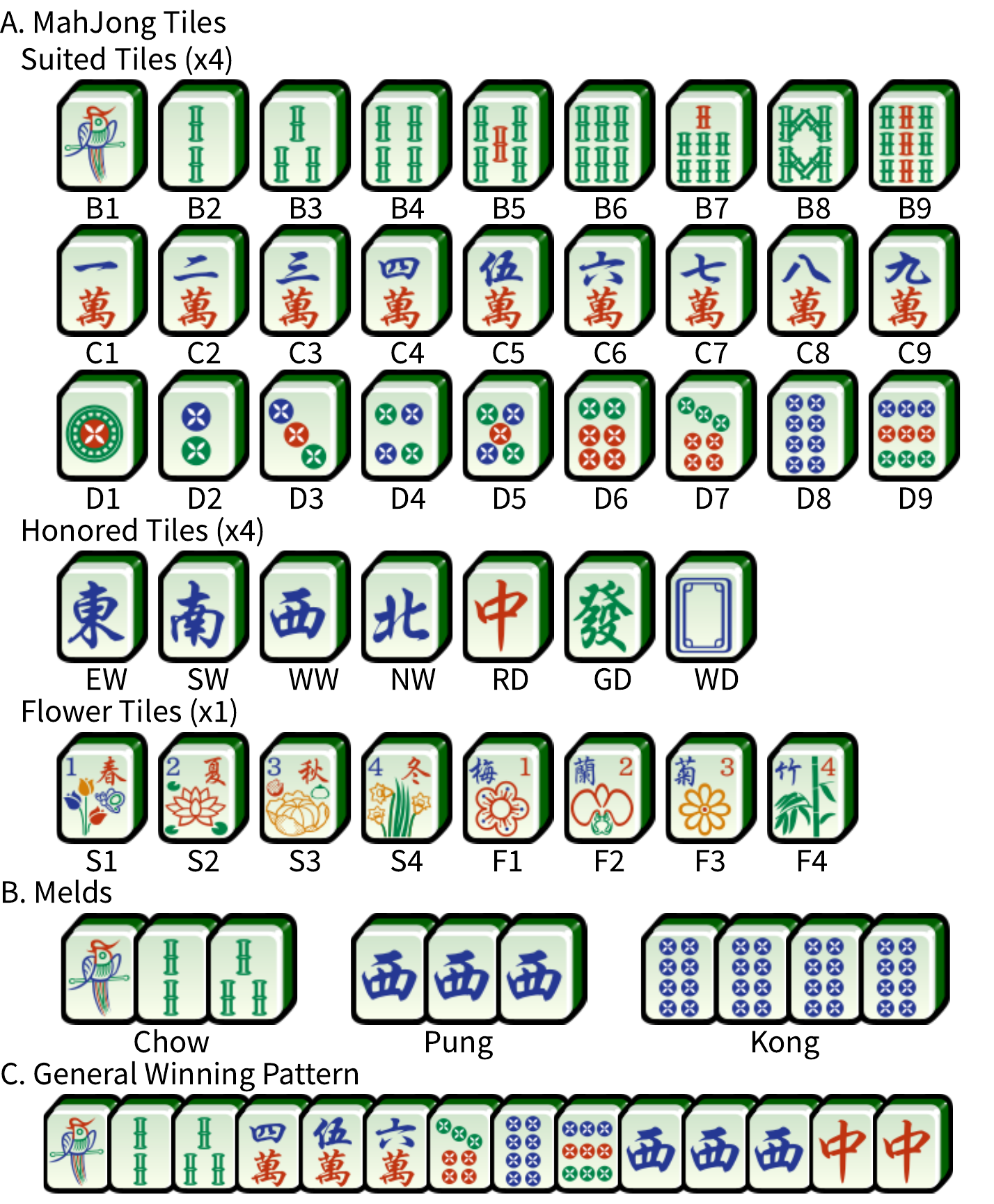} 
\caption{{Basics} 
 of Mahjong. (\textbf{A}). All the Mahjong tiles. There are four identical copies for suited tiles and honored tiles, and~one copy for each flower tile. (\textbf{B}). Examples of Chow, Pung, and~Kong. Note that only suited tiles are available for Chow. (\textbf{C}). Example of the general winning~pattern.}
\label{fig2}
\end{figure}

In Mahjong, there are a maximum of 144 tiles, as~shown in Figure~\ref{fig2}A. Despite its plethora of rule variants, Mahjong's general rules are the same. On~a broad level, Mahjong is a pattern-matching game. Each player begins with 13 tiles only observable to themselves, and~they take turns to draw and discard 1 tile until one completes a game goal with a 14th tile. The~general pattern of 14 tiles is four \textbf{melds} and a pair, as~shown in Figure~\ref{fig2}C. A~meld can take the form of \textbf{Chow}, \textbf{Pung}, and~\textbf{Kong}, as~shown in Figure~\ref{fig2}B. Apart from drawing all the tiles by themselves, players can take the tile just discarded by another player instead of drawing one to form a \textbf{meld}, called \textbf{melding}, or~declare a~win.

\subsection*{{Official} 
 International~Mahjong}

Official International Mahjong stipulates Mahjong Competition Rules (MCRs) to enhance the game's complexity and competitiveness and weaken its gambling nature. It specifies 80 scoring patterns with different points, called ``\textbf{fan}'', ranging from 1 to 88 points. In~addition to the winning patterns of four melds and a pair, players must have at least \mbox{eight points} by matching multiple scoring patterns to declare a~win. 

Specific rules and requirements for each pattern can be found in ~\cite{mcrEN}. Of~\mbox{81 \textbf{fans}}, 56 are highly valued and called \textbf{major fans} since most winning hands usually consist of at least one. The~standard strategy of MCR players is to make game plans by identifying several \textbf{major fans} closest to their initial set of tiles. Then, depending on their incoming tiles, they gradually select one of them as the terminal goal and strive to collect all remaining tiles before others do. The~exact rules of MCR are detailed in \textit{{Official} International Mahjong: A New Playground for AI Research}~\cite{a16050235}.

\section{Methods}

We first introduce the general concepts of Mxplainer and then present the details of the implementations in the following subsections. To~facilitate readability, we use uppercase symbols to indicate concepts, and~lowercase symbols with subscripts indicate  instances of~concepts.

\begin{figure}[H]
\includegraphics[width=0.98\textwidth]{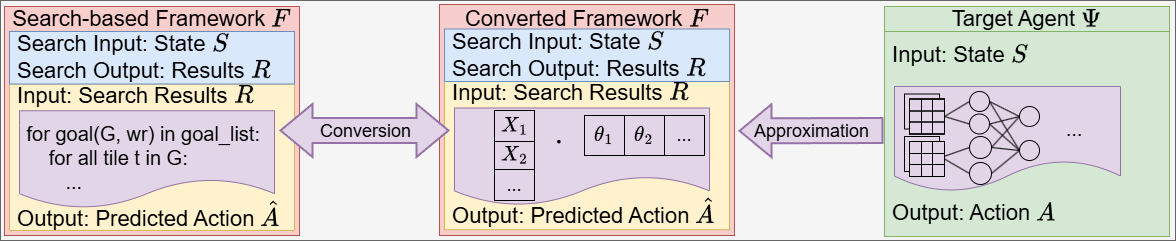} 
\caption{An overview of Mxplainer. Search-based framework $F$ is a manually engineered, domain-specific, parameterized template. A~component of $F$ can be converted into an equivalent network for supervised learning, where parameters of $F$ serve as neurons. Target agents $\Psi$ are black-box agents to be analyzed, and~they can be approximated by the converted $F$. }
\label{fig0}
\end{figure}

Figure~\ref{fig0} presents the concept overview of Mxplainer. We assume that the search-based nature of $F$ is explainable to experts in the domain, such as Mahjong players and researchers. Within~$F$, there are parameters $\Theta$ that control the behaviors of $F$. To~explain a specific target agent $\Psi$, we would like the behaviors of $F$ to approximate those of $\Psi$ as closely as possible. In~order to automate and speed up the approximation process, we convert a part of $F$ that contains $\Theta$ into an equivalent neural network representation, and~we leverage supervised learning to achieve the~goals.

\begin{figure}[H]
\includegraphics[width=0.65\columnwidth]{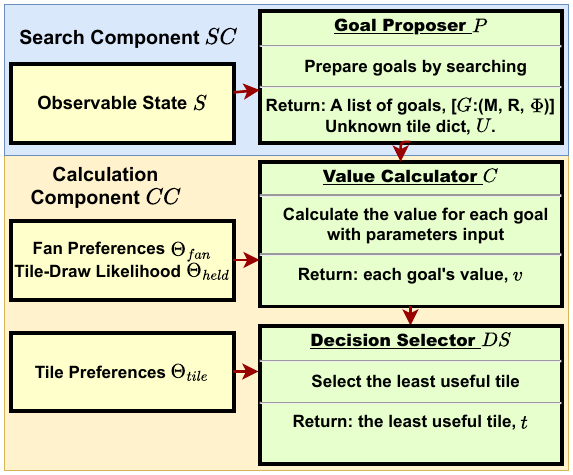} 
\caption{Components of framework $F$. Search Component $SC$ uses Goal Proposer $P$ for goal search. Calculation Component $CC$ consists of Value Calculator $C$ and Decision Selector $DS$, and~it is responsible for action calculations. The~least useful tile $t$ can be considered as the tile that appears least frequently among the most probable goals. Consequently, removing this tile has the least impact on the expected value.
}
\label{fig1}
\end{figure}

$F$ consists of Search Component $SC$ and Calculation Component $CC$, denoted as $F=SC|CC$. $SC$ searches and proposes valid goals in a fixed order. Next, $CC$ takes groups of manually defined parameters $\Theta$, each of which carries meanings and is explainable to experts, and~makes decisions based on the search results from $SC$, as~shown in Figure~\ref{fig1}.

$CC$ can be converted into an equivalent neural network $N$, for which its neurons are semantically equivalent to $\Theta$. $SC|N$ can approximate any target agents $\Psi$ by fitting $\Psi$'s state--action pairs. Since $\Theta$ is the same for both $CC$ and $N$, learned $\hat{\Theta}$ can be put back into $CC$ and explains actions locally through step-by-step deductions of $SC|CC$. Moreover, by~normalizing and studying $\Theta$, Mxplainer is able to compare and analyze agents' strategies and~characteristics. 

The construction of $F$ and the design of parameters $\Theta$ are problem-related, as~they reflect researchers' and players' attention to the game's characteristics and the game agents' strategies. The~conversion from $CC$ to $N$ and the approximation through supervised learning is problem-agnostic and can be automated. In~the Mxplainer framework, the~Search Component ($SC$) of $F$ is fixed and shared across all agents, while the behavior of the Calculation Component ($CC$) and its neural equivalent $N$ is determined by the parameters $\Theta$, which are tuned to mimic different target agents $\Psi$.

\subsection{Parameters $\Theta$ of Framework $F$}
For $\Theta$, we manually craft three groups of parameters to model people's general understanding of Mahjong: $\Theta_{tile}$, $\Theta_{fan}$, and~$\Theta_{held}$.

$\Theta_{tile}$ is used to break ties between tiles, and~different players may have different preferences for~tiles.

$\Theta_{fan}$ is designed to break ties between goals when multiple goals are equally distant from the current hand. There are more than billions of possible goals in Mahjong, but~each goal consists of \textbf{fans}. Thus, we use the compound of preferences of \textbf{fans} to sort goals. We hypothesize that there exists a natural order of difficulty between \textbf{fans}, which implies that some are more likely to achieve than others~\cite{mcrEN}. However, such an order is impossible to obtain unless there is an oracle that always gives the best action. Additionally, players break ties based on their inclinations towards \textbf{fans}. Since the natural difficulty order and players' inclinations are hard to decouple, we use $\Theta_{fan}$ to represent their products. Consequently, $\Theta_{fan}$ between different \textbf{fans} for a single player cannot be compared directly, but~$\Theta_{fan}$ for the same \textbf{fans} between different players reflect their comparative~preferences. 

$\Theta_{held}$ denotes the linear weights of a heuristic function that approximates the probability that a tile is held by other players. The~linear function takes features from local game states, including the number of total unshown tiles, the~length of game steps, and~the unshown tile counts of the neighboring tiles. Unshown tiles are those not visible to the agent, encompassing tiles in the wall and in opponents' hands. The~agent maintains a probability distribution over these tiles, modeled as a dictionary $U$, where each key is a tile type and its value is the estimated number of remaining copies. The~components and the usage of $\Theta_{held}$ will be discussed in detail in the following~sections. 

\subsection{Search Component $SC$ of Framework $F$}

In Mahjong, \textbf{Redundant Tiles}, $R$, refer to the tiles in a player’s hand that are considered useless for achieving their game goals. In~contrast, \textbf{Missing Tiles}, $M$, are the ones that players need to acquire in future rounds to complete their objectives. Following this, the \textbf{Shanten Distance} is defined as $D = |M|$, which conveys the distance between the current hand and a selected~goal. 

Here, we define a game goal as $G = (M, R, \Phi)$, since when both $M$ and $R$ are fixed, a~game goal is set, and~its corresponding \textbf{fans} $\Phi$ can be calculated. Each tile $t \in M$ additionally has two indicators, $i_p$ and $i_c$, which determine if it can be acquired through melding from other players. $i_p$ is true if the player already owns two tiles in the \textbf{Pung}, and~the same happens to $i_c$ and \textbf{Chow}. Thus, $ \forall t\in M$, it has $(t, i_p, i_c)$. 

Through dynamic programming, we can efficiently propose different combinations of tiles and test if they satisfy MCR's winning condition. We can search all possible goals, but~not all goals are reasonable to be considered as candidates. In~practice, only up to 64 goals, $G$, are returned in ascending \textbf{Shanten Distances} $D$, since in most cases, only the closest goals are important for decisions, and~they are usually less than two~dozen.

However, only the presence of goals is not enough. MCR players also need to consider other observable game information to jointly evaluate the value of each goal $G$, such as other players' discard history and unshown tiles. For~our framework $F$, we only consider unshown tiles $U$, a~dictionary keeping track of the number of each tile that is {not shown} 
 through observable~information.

Thus, the~Goal Proposer $P$ accepts game state information $S$ and~outputs $([G],U)$ such that $0<|[G]|\le64$, as~shown in Figure~\ref{fig1}A. In~most cases, goals, $G$, in $[G]$ can be split into several groups. Different groups contain different \textbf{fans} $\Phi$ and represent different paths to win, while the goals within each group have different tile combinations for the same \textbf{fans} $\Phi$.

\subsection{Calculation Component $CC$ of Framework $F$}

Calculation Component $CC$ of Framework $F$ consists of Value Calculator $C$ and Decision Selector $DS$. $CC$ contains three groups of tunable parameters $\Theta_{fan}$, $\Theta_{held}$, and~$\Theta_{tile}$ that control the behavior of Value Calculator $C$ and Decision Selector $DS$. In~all, these three groups of parameters, $\Theta_{fan}$, $\Theta_{held}$, and~$\Theta_{tile}$, are similar to parameters of neural networks; they are the key factors that control the behaviors of agents derived from framework $F$.

The Value Calculator $C$ takes the results of the Goal Proposer $P$, $([G], U)$ as input and estimates the value for each goal $G$. The~detailed algorithm of $C$ is shown in Algorithm~\ref{alg:algorithm1}. Simply put, $C$ multiplies the probability of successfully collecting each \textbf{missing tile} of a proposed goal as its estimated~value. 



The value estimation in the Value Calculator $C$ is governed by two parameter groups: $\Theta_{fan}$ and $\Theta_{held}$. The~probability of acquiring a required tile $t$ ({$t$ denotes a generic tile, and~$t \in X$ indicates that tile $t$ belongs to a set $X$}) 
 is decomposed into two components: drawing it from the wall or~\textbf{melding} it from another player’s~discard.

The probability of drawing tile $t$ is modeled as follows: 
$$P_{draw}(t) = U[t]/\Sigma(U)\cdot Z\cdot\Theta_{held}$$

This formulation reflects the assumption that the likelihood of drawing $t$ is proportional to the number of its remaining unshown copies $U[t]$ and~inversely proportional to the total number of unshown tiles $\Sigma(U)$. Furthermore, the~term $Z\cdot\Theta_{held}$ accounts for the effect of other players potentially holding the tile, where $Z$ denotes a feature vector derived from local game state attributes, and~$\Theta_{held}$ serves as a set of linear weights:
$$Z:\{\Sigma(U), \frac{1}{\Sigma(U)}, 1-\frac{1}{\Sigma(U)}, L, \frac{1}{L}, 1-\frac{1}{L}, U[t-2]:U[t+2], bias \}$$
\vspace{-15pt}
\begin{algorithm}[H]
\small
\caption{Value Estimation for A Single~Goal}
\label{alg:algorithm1}
\textbf{Input}: Goal $G$: $<$ \textbf{missing tiles} $M$:$ <t$, \textbf{Chow} indicator $i_{c}$, \textbf{Pung} indicator $i_{p}$$>$, \textbf{fan} list $F>$, unshown dict $U$, game length $L$\\
\textbf{Parameter}: $\Theta_{fan}$, $\Theta_{held}$\\
\textbf{Output}: Estimated value for goal G\\\vspace{-6pt}
\begin{algorithmic}[1]
\STATE Initialize  value $v \leftarrow 100$
\FORALL {\textbf{missing tile} $m \in M $}
\STATE // Construct local tile feature $x_m$ from game length $L$,\\
\STATE // and remaining adjacent tile count $U$.\\
\STATE $x_m \leftarrow U, L $ \\
\STATE // Calculate probability of drawing $m$ and others discarding $m$
\STATE $p_{draw} \leftarrow U[m]/sum(U)* \Theta_{held}*x_m$ \\
\STATE $p_{discard} \leftarrow U[m]/sum(U)$  \\
\STATE $source \leftarrow 0$
\IF {$i_p$}
\STATE $source \leftarrow 3$ // one can pung from all others
\ELSIF {$i_c$}
\STATE $source \leftarrow 1$ // one can only \textbf{chow} from the one left
\ENDIF
\STATE $p_{meld} \leftarrow p_{discard} * source$
\STATE $v \leftarrow v \times (p_{draw} + p_{meld})$
\ENDFOR
\STATE Total \textbf{fan} weight $fw \leftarrow \sum {\Theta_{fan}[f]}$ for \textbf{fan} $f \in F$
\STATE $v \leftarrow v \times fw$

\STATE \textbf{return} wr
\end{algorithmic}
\end{algorithm}

Here, $L$ denotes the current game length, and~$U[t-2]:U[t+2]$ represents the unshown counts of tiles adjacent to $t$. Including a bias term, $\Theta_{held}$ contains 12 learnable weights in total. In~contrast, the~probability of \textbf{melding} tile $t$ from another player’s discard is modeled using a uniform distribution, $P_{meld}(t) = U[t]/\Sigma(U)$. We examine alternative probability models for $P_{draw}$ and $P_{meld}$ in Section~\ref{sec6.2}, and~we further discuss the rationale behind our modeling choices in Section~\ref{sec7.1}.

The Decision Selector $DS$ collects results from the Value Calculator $C$ and calculates the final action based on each goal's estimated value. The~detailed algorithm for $S$ is shown in Algorithm~\ref{alg:algorithm2}. The~goal of $S$ is to efficiently select the tile to discard with the most negligible impact on the overall value. Heuristically, the~required tiles of goals with higher values are more important than those with lower ones. Conversely, the~\textbf{redundant tiles} of such goals are more worthless since their existence actually hinders the goals' completion. Thus, each tile's worthless degree can be computed by accumulating the value of goals that regard it as a \textbf{redundant tile}, and~the tile with the highest worthless degree has the most negligible impact on the overall value. Decision Selector $DS$ accepts $\Theta_{tile}$ as parameters, which are used similarly as $\Theta_{fan}$ to break ties between tiles. For~action predictions, such as \textbf{Chow} or \textbf{Pung}, $F$ records values computed by $C$, assuming those actions are taken, and~$F$ selects the action with the highest~value.

\begin{algorithm}[H]
\small
\caption{Discarding Tile~Selection}
\label{alg:algorithm2}
\textbf{Input}: List of goals and their values $L=[($Goal $G$, value $v)]$, Hand tiles $H$\\
\textbf{Parameter}: $\Theta_{tile}$\\
\textbf{Output}: Tile to discard\\\vspace{-6pt}
\begin{algorithmic}[1]
\STATE Initialize tile values $d \leftarrow$ Dict $\{$tile $t:0\}$
\FORALL{$(G, v) \in L$}
\FORALL{tile $t \in G$}
\STATE $d[t] \leftarrow d[t] + v$ // \textbf{redundant tiles}
\ENDFOR
\ENDFOR
\FORALL{tile $t \in d$}
\STATE $d[t] \leftarrow d[t] \times \Theta_{tile}[t]$
\ENDFOR
\STATE \textbf{return} $\arg\max_{t \in H}{d[t]}$
\end{algorithmic}
\end{algorithm}

\subsection{Differentiable Network N of~Mxplainer}

The search-based Framework $F$ is a parameterized search-based agent template, and~its parameter $\Theta$ needs to be tuned to approximate any target agents' behaviors. Luckily, Calculator $C$ and Decision Selector $DS$ only contain fixed-limit loops, with~each iteration independent from others and if--else statements whose outcomes can be pre-computed in advance. Thus, $C$ and $S$ can be converted into an equivalent neural network $N$ for parallel computation, and~the parameters $\Theta$ can be optimized through batched gradient descent. The~rules for the conversion can be found in {Appendix} 
 \ref{secB}.
\begin{equation}
  \mathbb{I} =
    \begin{cases}
      1 & \text{if it is real data}\\
      0 & \text{otherwise}
    \end{cases}   
\label{eqn:1}
\end{equation}

To distinguish between padding and actual values, we define a value indicator as Equation~\eqref{eqn:1} to facilitate parallel computations. Then, Algorithm~\ref{alg:algorithm1} and Algorithm~\ref{alg:algorithm2} can be easily re-written as Listing~\ref{lst:alg1} and Listing~\ref{lst:alg2}. The~inputs to the Listings~are as follows: 
\begin{itemize}
  \item $F\in \mathbb{R}^{80}$, $U \in \mathbb{N}^{34}$, $H\in \mathbb{N}^{64\times34}$, $X\in \mathbb{R}^{34\times12}$: Vectorized $F$, $U$, $H$, and~$x_m$ from Algorithm~\ref{alg:algorithm1} and Algorithm~\ref{alg:algorithm2}.
  \item $MT:(\mathbb{I}, M) \in \mathbb{R}^{34\times2}$: \textbf{Missing tiles} $M$ with indicators.
  \item $BM\in \mathbb{N}^{34\times3}$: One-hot branching mask from $i_{c}$ and $i_{p}$.
  \item $V \in \mathbb{R}^{64\times1}$: The computed value from Listing~\ref{lst:alg1}.
  \item w\_held, w\_fan, w\_tile: $\Theta_{held}$, $\Theta_{fan}$, and~$\Theta_{tile}$.
\end{itemize}
\vspace{-6pt}
\begin{listing}[H]%
\caption{Paralleled Algorithm~\ref{alg:algorithm1} in PyTorch 2.3.1}%
\label{lst:alg1}%
\begin{lstlisting}[language=Python,xleftmargin=2.2em]
def calc_winrate(F, MT, BM, X, U, w_held, w_fan):
    v = 100
    p_uniform = MT[:, 1]* U/torch.sum(U)
    p_draw = p_uniform*X*w_held
    p_discard = p_uniform
    # encode chi, peng, no-op coeff
    cp_cf = torch.tensor([3, 1, 0])
    # compute tile-wise cp_coeff
    cp_cf = cp_cf * BM
    cp_cf = torch.sum(cp_cf, dim=1)
    p_meld = p_discard*cp_cf
    p_tile = p_darw+p_meld
    # use indicator to set paddings 0
    masked_tile = MT[:, 0]*p_tile
    # convert 0's to 1 for product op
    masked_tile += 1-MT[:, 0]
    v *= torch.prod(masked_tile, dim=0)
    #calculate preference for fan
    p_fan = F*w_fan
    return v*p_fan
    
\end{lstlisting}
\end{listing}

\begin{listing}[H]%
\caption{Paralleled Algorithm~\ref{alg:algorithm2} in PyTorch}%
\label{lst:alg2}%
\begin{lstlisting}[language=Python]
def sel_tile(H, V, w_tile):
    tile_v = H*V
    tile_v = torch.sum(tile_v, dim=0)
    tile_pref = tile_v*w_tile
    return torch.argmax(tile_pref)
    
\end{lstlisting}
\end{listing}

\subsubsection*{{The} 
 Resulting Network N and the Training~Objective}

The sizes and meanings of the network $N$'s outputs and three groups of parameters are reported in Table~\ref{tab:network_size}. Action Kong is not explicitly encoded in the action space. Instead, it is logically implied and automatically executed by the Mxplainer agent under the following~conditions:
\begin{itemize}
\item \textbf{Converting a Melded Pung:} The agent has an existing, exposed \textbf{Pung} of tile $t$ and tries to discard the fourth $t$. The~agent then executes \textbf{Kong} automatically.
\item \textbf{Forming a Concealed Kong:} The agent has three identical tiles $t$ concealed in its hand. \textbf{Kong} is triggered when it draws the fourth $t$ and tries to discard it.
\item \textbf{Forming a Kong:} The agent has three identical tiles $t$ concealed in its hand. \textbf{Kong} is triggered automatically, when another player plays tile $t$, and~the agent tries to \textbf{Pung} and discard $t$ at the same time.
\end{itemize}

\vspace{-6pt}

\begin{table}[H]
\caption{Sizes and meanings of neural network $N$'s~components.}
\begin{tabularx}{\textwidth}{LLl}
\toprule
     \textbf{Entry} & \textbf{Size} & \textbf{Meaning} \\
\midrule
    Overall Output& 39 & Combined output size \\
    Output (Action) & 5 & 5 Actions:  \\
     &  & Pass, 3 types of \textbf{{Chow}
}, \textbf{{Pung}
} \\
    \midrule
    Param $\Theta_{fan}$ & 80 &	Preference for all 80 \textbf{fans}  \\
    Params $\Theta_{held}$ & 12 &	Linear features weights w/bias  \\
    Param $\Theta_{tile}$ & 34 & Preference for all 34 tiles  \\

\bottomrule
\end{tabularx}

\label{tab:network_size}
\end{table}

In the framework's decision process, declaring \textbf{Kong} in these scenarios is always the optimal action, as~it increases the hand's value without altering the strategic plan. This design formally decomposes the policy into a piecewise-linear component for high-level actions and a domain-logic component for specialized decisions like discards and Kongs. The~resulting composition can be viewed as building a complex policy from simpler, interpretable sub-policies, a~concept with rigorous footing in optimization theory~\cite{10.1007/978-1-4614-1927-3_10, Griewank01122013, 11195691}.

The learning objectives depend on the form of target agents $\Psi$ and the problem context. Since we are modeling the selection of actions as classification problems, we use cross-entropy (CE) loss between the output of $N$ and the label $Y$. The~label can be soft or hard depending on whether the target agent gives probability distributions. Since we cannot access action distributions from Botzone's game log, we use actions as hard ground-truth labels for all target agents. Additionally, an~L2-regularization term of $(\Theta_{fan}-abs(\Theta_{fan}))^2$ is added to penalize negative values of \textbf{fan} preferences to make the heuristic parameters more~reasonable.

We employ supervised learning for this imitation task as it provides a direct and stable optimization objective to achieve high behavioral fidelity, which is a prerequisite for generating faithful~explanations.

After training, the~learned parameters $\Theta$ can be directly filled back into $CC$. Since $CC$ is equivalent to $N$, $SC|CC(\Theta)$ inherits the characteristics of $SC|N(\Theta)$, approximating the target agent $\Psi$. By~analyzing the parameterized agent $SC|CC(\Theta)$, we can study the parameters to learn the comparative characteristics between agents and gain insights into the deduction process of $\Psi$ through the step-by-step algorithms of Framework $SC|CC$. Since only data pairs are required from the target agents, we can use Mxplainer on both target AI agents and human~players. 

\section{Experiments}

We conducted a series of experiments to evaluate the effectiveness of Mxplainer in generating interpretable models and analyze their explainability. Three different target agents are used in these experiments. The~first agent $\psi_1$ is a search-based MCR agent with manually specified characteristics as a baseline. Specifically, this agent only considers the \textbf{fan} of \textit{Seven Pairs} to choose, and~all its actions are designed to reach this target as efficiently as possible. When multiple tiles are equally good to discard, a~fixed order is predefined to break the equality. The~second agent $\psi_2$ is the strongest MCR AI from an online game AI platform, Botzone~\cite{10.1145/3197091.3197099}. The~third agent $\psi_3$ is a human MCR player from an online Mahjong platform, MahjongSoft. 

Around 8000 and 50,000 games of self-play data are generated for the two AI agents. Around 34,000 games of publicly available data are collected from the game website for the single human player over more than 3 years. The~datasets for each agent were split into training, validation, and~test sets on a per-game basis to prevent data leakage and ensure a robust evaluation. Through supervised learning on these datasets, three sets of parameters $\theta_{1,2,3}$ are learned and filled back in the search-based framework $F$ as-is, creating interpretable white-box agents $\hat{\psi}_{1,2,3}$ with similar behavior to the target agents $\psi_{1,2,3}$. The~weights for each agent are selected with the highest validation accuracy from three runs. To~make weights comparable across different agents, reported weights are normalized with $100\times(\Theta-\Theta_{min})/\Sigma(\Theta-\Theta_{min})$.

Since it is common in MCR where multiple \textbf{redundant tiles} can be discarded in any order, it is difficult to achieve high top-1 accuracy on the validation sets, especially for $\psi_2$ and $\psi_3$, which have no preference on the order of tiles to discard. As~a reference of the similarity, $\hat{\psi}_{1,2,3}$ achieves the top-three accuracy of 97.15\%, 93.47\%, 90.12\% on the data of $\psi_{1,2,3}$ ({All the accuracy figures presented in this paper only take into account states $\ge2$ valid actions, which account for less than 30\% of all states. The~overall accuracy can be roughly calculated as $70+0.3\times acc$, where $acc$ represents the reported accuracy.}).

\subsection{Correlation Between Behaviors and~Parameters}

In the search-based framework $F$, the~parameters, $\Theta$, are designed to be strategy-related features that should take different values for different target agents. Here, we analyze the correlation between the learned parameters and target agents' behavior to prove that Mxplainer can extract high-level characteristics of agents and explain their~behavior.

\subsubsection{{Preference} 
 of \textbf{Fans} to Choose}

$\Theta_{fan}$ stands for the relative preference of agents to win by choosing each \textbf{fan}. For~the baseline target $\psi_1$, which only chooses the \textbf{fan} of \textit{Seven Pairs}, the~top values of $\theta_{1,fan}$ learned are shown in Table~\ref{tab:agent23_fan}B. We also include $\theta_{2,fan}$ values for comparisons, demonstrating the differences in learned weights between specialized \textit{Seven Pairs} agent and other agents. It can be seen that the weight for \textit{Seven Pairs} is much higher than those of other \textbf{fans}, showing the strong preference of $\hat{\psi}_1$ to this \textbf{fan}. The~\textbf{fans} with the second and third largest weights are patterns similar to \textit{Seven Pairs}, indicating that $\hat{\psi}_1$ also tends to approach them during the gameplay based on its learned action choices, though~it eventually ends with \textit{Seven Pairs} for 100\% of the games. For~targets $\psi_2$ and $\psi_3$ with unknown preferences of \textbf{fans}, we count the frequency of occurrences of each \textbf{major fan} in their winning hands as an implication of their preferences and compare these two agents on both the frequency and the learned weights of each \textbf{fan}. Table~\ref{tab:agent23_fan}C shows all the \textbf{major fans} with a difference of 1\% in frequency. Except~for the last row, the~data shows a significant positive correlation between the preference of target agents $\psi$ and the learned $\theta_{fan}$. 

\begin{table}[H]
\caption{(A). Defined order for tiles and their learned weights. (B). Sorted weights in the learned parameters $\theta_{1,fan}$ and their corresponding \textbf{fans}. Learned weights for $\theta_{2,fan}$ are added for comparison.
(C). The~\textbf{major fans} of $\psi_2$ and $\psi_3$ with a difference of at least 1\% in historical frequency, which is calculated from their game history {data.} 
}
\begin{tabularx}{\textwidth}{rrCC}
\toprule
     A& Tile Name & Defined Order & $\theta_{1,tile}$  \\
\midrule
     & White Dragon & 1 & 1.41\\
     & Green Dragon & 2 & 1.13\\
     & Red Dragon & 3 & 1.07\\
\midrule
     B & \textbf{Fan} Name & $\theta_{1,fan}$ & $\theta_{2,fan}$\\
\midrule
     & \textit{{Seven Pairs} 
} &	56.19 & 3.59\\
     & \textit{{Pure Terminal Chows} 
} &	7.42 & 1.99\\
     & \textit{{Four Pure Shift Pungs} 
} &	4.10 & 0.42\\
\midrule
 C&\textbf{Fan} Name & Frequency Diff & Param Diff \\ 
 && $\psi_2 - \psi_3$ (\%)  & $\theta_{2,fan} - \theta_{3,fan}$ \\ 
\midrule
&\textit{{Pure Straight} 
} & 1.30  & 0.45 \\
&\textit{{Mixed Straight} 
} & 1.94  & 0.65 \\
&\textit{{Mixed Triple Chow}  
} & 2.09  & 0.63 \\
&\textit{{Half Flush} 
} & 1.04 & 0.27 \\
&\textit{{Mixed Shifted Chows} 
} & 5.08  & 0.31 \\
&\textit{{All Types} 
} & 2.60  & 0.27 \\
&\textit{{Melded Hand} 
} & $-$8.25 & $-$1.24 \\
&\textit{{Fully Concealed Hand} 
} & 1.83  & $-$0.55 \\ 
\bottomrule
\end{tabularx}

\label{tab:agent23_fan}
\end{table}

\subsubsection{{Preference} of Tiles to Discard}

$\Theta_{tile}$ stands for the relative preference of discarding each tile, especially when multiple \textbf{redundant tiles} are valued similarly. Since $\psi_2$ and $\psi_3$ show no apparent preference in discarding tiles, we focus on the analysis of $\psi_1$, which is constructed with a fixed tile preference. We find the learned weights almost form a monotonic sequence with only 3 exceptions out of 34 tiles, showing strong correlation between the learned parameters and the tile preferences of the target agent. Table~\ref{tab:agent23_fan}A shows the first a few entries of $\theta_{1,tile}$.

\subsection{Manipulation of Behaviors by~Parameters}
Previous experiments have shown that greater frequencies of \textbf{fans} within the game's historical record lead to elevated preferences. In~this experiment, we illustrate that through the artificial augmentation of \textbf{fan} preferences, the~modified agents, in~contrast, display elevated frequencies of the corresponding \textbf{fans}. We make adjustments to the parameters $\theta_{2,fan}$ within $\hat{\psi}_2$ by multiplying the weight assigned to the \textit{All Types} fan by a factor of 10 ({The factor of 10 is an arbitrary but significant multiplier chosen to induce a strong and clearly observable shift in the agent's behavior, thereby demonstrating that the learned parameters in $\Theta_{fan}$ are actionable and directly influence strategic preferences.}). This gives rise to the generation of a new agent, $\psi'_2$, which is expected to exhibit a stronger inclination towards selecting this \textbf{fan}.

We collect roughly around 8000 self-play game data samples from $\psi_2$ and $\psi'_2$. Subsequently, we determine the frequency with which each fan shows up in their winning hands. The~data indicates that among all the fans, only the \textit{All Types} fan undergoes a frequency variation that surpasses 1\%. Its frequency has risen from 2.59\% to 5.76\%, signifying an increment of 3.17\%. In~contrast, the~frequencies of all other \textbf{major fans} have experienced changes of less than 1\%. Given that Mahjong is a game characterized by a high level of randomness in both the tile-dealing and tile-drawing processes, adjusting the parameters related to fan preferences can only bring about a reasonable alteration in the actual behavior patterns of the target agents. In~conjunction with the findings of previous experiments, we can draw the conclusion that the parameters within Mxplainer are, in~fact, significantly correlated with the behaviors of agents. Moreover, through the analysis of these parameters, we are capable of discerning the agents' preferences and high-level behavioral~characteristics.

\subsection{Interpretation of Deduction~Process}

In this subsection, we analyze the deduction process of the search-based framework $F$ on an example game state by tracing the intermediate results of $\hat{\psi}_2$ to demonstrate the local explainability of Mxplainer agents in~decision-making.

The selected game state $S$ is at the beginning of a game with no tiles discarded yet, and~the player needs to choose one to discard, as~shown in Table~\ref{tab:sample_gamestate}A. The~target black-box agent $\psi_2$, selects the tile B9 as the optimal choice to discard for unknown reasons. However, analyses of the execution of the white-box agent $\hat{\psi}_2$ explain the~choice.

\begin{table}[H]
\caption{(A). An~example game state at the beginning of the game where no
tiles have been discarded by other players. (B). Some proposed goals for state $s$ by $\hat{\psi}_2$ and the estimated win probabilities. ``H\&K'' is an abbreviation for \textit{Honors and Knitted} due to space constraints.}
\begin{adjustwidth}{-\extralength}{0cm}
\centering
		\begin{tabularx}{\fulllength}{lllllL}
\toprule
 A&& Other Information &  &   & Current Hand\\ 
\midrule
&&None&&&{\includegraphics[width=5cm, height=0.5cm]{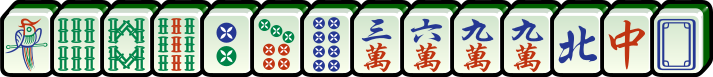}} \\
\midrule
B&ID & \textbf{Major Fan}& Probs & \textbf{Redundant Tiles} $R$  & Final Target\\ 
\midrule
&25 & \textit{Lesser H\&K Tiles}~\textsuperscript{1} & 1.000 & D7, C9, B6, B8, \underline{B9}  &{\includegraphics[width=5cm, height=0.5cm]{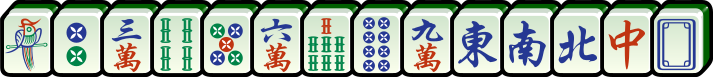}}\\
&0 & \textit{Knitted Straight}~\textsuperscript{1} & 0.125 & C9, \underline{B9}, NW, RD, WD  &{\includegraphics[width=5cm, height=0.5cm]{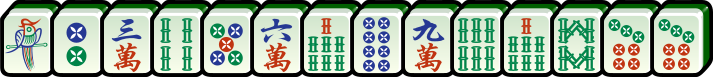}}\\
&11 & \textit{Pure Straight} & 0.018 & D2, C3, C6, NW, RD, WD  &{\includegraphics[width=5cm, height=0.5cm]{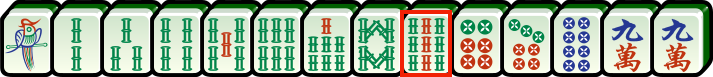}}\\
&50 & Seven Pairs~\textsuperscript{1} & 0.017 & B6, B8, \underline{B9}, NW, RD, WD  &{\includegraphics[width=5cm, height=0.5cm]{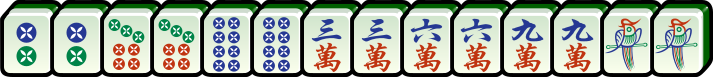}}\\
\bottomrule
\end{tabularx}
\end{adjustwidth}
	\noindent{\footnotesize{\textsuperscript{1} A \textbf{fan} that does not follow the pattern of four melds and a pair.}}

\label{tab:sample_gamestate}
\end{table}

The Search Component $SC$ proposed 64 possible goals for $s$, and~Table~\ref{tab:sample_gamestate}B shows a few goals representative of different \textbf{major fans}. With~fitted $\theta_2$, Algorithm~\ref{alg:algorithm1} produces an estimated value for each goal $G$, as~listed under the ``Value'' column of Table~\ref{tab:sample_gamestate}B. We observe that though B9 is required in goals such as \textit{Pure Straight}, it is not required for many goals with higher estimated values, such as \textit{Knitted Straight}.

Following Algorithm~\ref{alg:algorithm2}, we accumulate the values for tiles in \textbf{Redundant Tiles} $R$. The~higher value of a tile indicates a higher value if the tile is discarded, and~B9 turns out to have a higher value than other tiles by a large margin, which is consistent with the observed decision of $\psi_2$.

This section merely demonstrates an analysis of action selection within the context of a simple Mahjong state. However, such analyses can be readily extended to other complex states. The~Search Component $SC$ consistently takes in information and puts forward the top 64 reachable goals, ranked according to distances. The~Calculation Component $CC$ calculates the value for each goal using the learned parameters. Finally, an~action is chosen based on these~values.

Without Mxplainer, people can only observe the actions of black-box MahJong agents, but~they cannot understand how the decisions are made. Our experiments show that Mxplainer's fitted parameters can well approximate and mimic target agents' behaviors. By~examining the fitted parameters and Mxplainer's calculation processes, experts are able to interpret the considerations of black-box agents leading to their~actions.

\section{Ablation and Comparative~Study}
\unskip
\subsection{Number of Searched~Goals}
To assess potential bias from the 64-goal cap, we performed a sensitivity analysis on goal recall for the target agent $\psi_2$. Given that the probability of success decreases exponentially with Shanten Distance $D$ (approximately by a factor of $1/34$ per unit increase), we specifically measured the recall rate for goals with the minimum $D$ in each~state.

A survey of approximately 10,000 game logs revealed that the average number of such minimum-$D$ goals was 13.28. The~coverage rate for these goals and the corresponding search time are summarized in Table~\ref{tab:ablation}.

\begin{figure}[H]
\includegraphics[width=0.65\columnwidth]{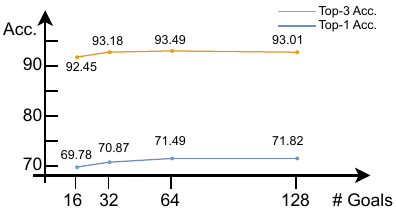} 
\caption{Top-1 and top-3 accuracy of $\hat{\psi_2}$ when the number of goals is 16/32/64/128.
}
\label{fig:ablation}
\end{figure}

We further evaluated the predictive accuracy of the approximated agent $\hat{\psi_2}$ under different goal caps (16, 32, 64, and~128). The~results, illustrated in Figure~\ref{fig:ablation}, show that both Top-1 and Top-3 accuracy improve substantially when the cap is increased from 16 to 64. However, increasing the cap further to 128 yields only marginal gains, indicating diminishing returns beyond 64~goals.

\begin{table}[H]
    \caption{Recall rate and time cost for goals with different~caps.}
\begin{tabularx}{\textwidth}{CCC}
    \toprule
        \textbf{Cap (K)} & \textbf{Coverage Rate@K} & \textbf{Time (s)} \\
        \midrule
        16 &  79.31\%&  0.097 \\
        32 &  87.85\%& 0.098 \\
        64 &  96.82\%&  0.109\\
        128 &  99.79\%&  0.128 \\
        \bottomrule
    \end{tabularx}
    \label{tab:ablation}
\end{table}
\vspace{-12pt}

\unskip

\subsection{Distribution~Modeling}\label{sec6.2}
To rigorously evaluate the impact of probability model specification, we conducted an ablation study comparing different configurations for modeling $p_{draw}$ and $p_{discard}$.

The results are summarized in Table~\ref{tab:distribution_ablation}. Here, ``Network'' denotes a learnable module that uses local game features $Z$ to estimate the acquisition probability, whereas ``Uniform'' assumes a simple uniform distribution over unshown tiles. The~configuration highlighted in bold was selected for our final framework, as~it achieved the highest overall accuracy without a substantial increase in parameter count. This suggests that accurately modeling the draw probability from the wall is more critical for policy imitation than refining the discard model under the current~framework.
\begin{table}[H]
    \caption{Ablation study on probability models for tile~acquisition. }
\begin{tabularx}{\textwidth}{CCCCC}
    \toprule
       \boldmath{$p_{draw}$} & \boldmath{$p_{discard}$} & \textbf{Top-1 Acc} & \textbf{Top-3 Acc } &\textbf{Parameter Size}\\
        \midrule
        Uniform &  Uniform& 66.26 & 90.65&114\\
        \textbf{{Network}
} &  \textbf{{Uniform} 
} &  \textbf{{71.49} 
} & \textbf{{93.47}
}& \textbf{{126} 
}\\
        Uniform &  Network& 67.35 & 91.16&126\\
        Network &  Network &  71.49 & 93.44&138\\
        \bottomrule
    \end{tabularx}
    \label{tab:distribution_ablation}
\end{table}

\subsection{Parameter Size and Methods of Behavior~Cloning}
In an effort to boost the explainability of Mxplainer, we have minimized the parameter size of the Small Network $N$. Nevertheless, this design might compromise its expressiveness and decrease the accuracy of the approximated agents. Consequently, we aim to further investigate the impact of the parameter size of the Small Network $N$ on the approximation of the Target Agent $\Psi$. Specifically, we augment the parameter size and train networks $\hat{\psi_2'}$ and $\hat{\psi_2''}$ to approximate the identical Target Agent $\psi_2$. We do not elaborate on the modifications to the network structures herein. The~parameter sizes and their respective top-3 accuracies are presented in Table~\ref{tab:network_comps}A.

We observe that by increasing the parameter size, Mxplainer can approximate the Target Agent $\Psi$ with greater accuracy. Nevertheless, the~drawback lies in the fact that the larger the number of parameters in the Small Network $N$, the~more challenging it becomes to explain the actions and the meaning of the parameters to users. Although~the approximated agents cannot replicate the exact actions of the original black-box agents, they can account for their actions in most scenarios, as~demonstrated by previous~experiments.

We also conduct a comparison of the effects of different methods. Specifically, we construct a random forest consisting of 100 decision trees and employ a pure neural network to learn the behavior policy of $\psi_2$ through behavior cloning. The~results are presented in Table~\ref{tab:network_comps}B. Although~decision trees are inherently self-explanatory, their low accuracies render them unsuitable for Mahjong, which involves numerous OOD states. The~accuracy of Mxplainer is not notably lower than that of traditional neural networks with sufficient expressivity, but~it has the advantage of being able to explain the reasoning underlying the~actions.

\begin{table}[H]
\caption{(A). Comparison between parameter size and resulting accuracies. (B). Comparison between different methods and~{accuracies.} 
 }
\begin{tabularx}{\textwidth}{lLllllL}
\toprule
      A &Network &Param& $\theta_{fan}$& $\theta_{held}$&$\theta_{tile}$ &Top-3 Acc\%\\
\midrule
    &$\hat{\psi_2}$ &126& 80&	12&34&93.47\\
    &$\hat{\psi_2'}$ &2804& 2720&	50&34&93.77\\
    &$\hat{\psi_2''}$ &3280& 2800&	435&45&94.57\\
\midrule
      B &Method &Param& &Acc\%& &Top-3 Acc\%\\
\midrule
    & {\small Mxplainer} & 126&&71.49&&93.47\\
    & {\small Decision Tree} & N/A&&28.91&&34.82\\
    & {\small Behavior Clone} & 34M&&75.31&&95.69\\
\bottomrule
\end{tabularx}
\label{tab:network_comps}
\end{table}
\unskip

\section{Discussion}

A primary limitation of Mxplainer is that its fidelity is inherently bounded by the expressiveness of the hand-crafted search framework F. Strategies that operate outside the paradigm of goal-search and linear parameterization may not be fully~captured.

\subsection{Choice of Probability Distributions for~Tiles}\label{sec7.1}
A key design choice in the Mxplainer framework is the use of different probability models for drawing a tile from the wall ($p_{draw}$) and for an opponent discarding a tile ($p_{discard}$). We model $p_{draw}$ with a learnable neural component that incorporates local game features, while $p_{discard}$ is modeled using a uniform distribution over the unshown~tiles.

This choice reflects a deliberate trade-off between expressiveness and interpretability. Modeling $p_{draw}$ with a learnable network is both feasible and critical. The~wall is a passive, stochastic element; its state can be reasonably approximated by tracking unshown tiles. A~model that learns to weight features like total unshown tiles, game length, and~availability of adjacent tiles ($Z$) can effectively capture the nuanced likelihood of drawing a specific tile, providing a significant accuracy boost over a uniform assumption, as~shown in our ablation study (Section \ref{sec6.2}).

In contrast, we employ a uniform distribution for $p_{discard}$. This assumption is a significant simplification, as~it likely introduces a systematic bias by overstating the melding probability for tiles that opponents would strategically withhold. In~theory, this could inflate the learned value of goals that rely on melding such~tiles.

However, our ablation study (Section \ref{sec6.2}) demonstrates that the core conclusions of Mxplainer, specifically, its high imitation accuracy and the strategic insights derived from $\Theta$, are remarkably robust to this simplification. The~minimal performance gap between uniform and learned discard models suggests that the framework successfully absorbs much of this potential bias into the other calibrated parameters (e.g., $\Theta_{fan}$) while maintaining a highly interpretable structure. Therefore, while a more sophisticated opponent model is a worthwhile goal for future work, the~uniform assumption is not a prerequisite for Mxplainer's current ability to deliver faithful and explainable agent imitations. We prioritized this simpler, interpretable model to maintain a clear causal chain of reasoning, accepting a theoretically sub-optimal but empirically adequate approximation for the purpose of high-fidelity policy~imitation.

Furthermore, the~imitation accuracy of Mxplainer approaches that of a large, black-box neural network trained via behavior cloning (see Table~\ref{tab:network_comps}B). This near-state-of-the-art performance, achieved with only 126 interpretable parameters compared to 34 million, underscores the efficiency of our parameterized search framework. It validates that the high fidelity of the approximation is not sacrificed for explainability but~rather achieved through a structured, interpretable~model.

\subsection{Compare Characteristics of Agents with~Mxplainer}
With Mxplainer, we can compare the differences between agents. By~comparing the fitted parameters in parallel with other agents, we can analyze the characteristics of different agents. For~example, we can easily observe that the black-box AI agent $\psi_2$ has a much higher weight on \textit{Thirteen Orphans}, \textit{Seven Pairs}, and~\textit{Lesser H\&K Tiles}. In~contrast, the~human player $\psi_3$ has a significant inclination of making \textit{Melded Hand}, and~these observations are indeed backed by their historical wins in Supplemental Material Table S2. Note that while Mxplainer successfully profiles an individual's strategy, its ability to capture population-level human playstyles requires future study with a broader~dataset.

\subsection{Limitations}
While Mxplainer provides faithful and interpretable approximations, its fidelity is inherently bounded by the expressiveness of the manually constructed search-based framework $F$. Strategies that operate on principles fundamentally beyond the template’s goal-search logic and linear parameterization—for instance, those relying on complex, non-linear state evaluations or long-term deceptive plays not captured by our goal-oriented model—may not be fully captured. This is a deliberate trade-off we make to prioritize interpretability, as~the components of $F$ are designed to be comprehensible to domain~experts.

\subsection{Extend Mxplainer to Other Settings and Future~Direction}
Our proposed approach has a unique advantage in quantifying and tuning weights for custom-defined task-related features in areas where interpretability and performance are crucial. While Mxplainer is specifically designed for Mahjong, we hypothesize that its unique approach and XAI techniques may be applicable to other applications. In~fact, we experimentally applied our proposed framework to two examples: Mountain Car from Gym~\cite{brockman2016openai} and Blackjack~\cite{10.5555/3312046}. Both examples, which can be found in {Appendix}
 \ref{secA}, confirmed the effectiveness of our proposed method. However, the~scope of application of our method still requires further~study. 

Another compelling future direction is to extend Mxplainer into a multimodal framework. By~incorporating auxiliary data streams—such as eye-tracking to measure attention, heart rate variability to assess cognitive load or stress, and~mouse-movement dynamics—we could move beyond modeling a static strategic policy. This would allow us to create a context-aware model that explains how a player's decisions are modulated by their real-time psychological and physiological state. Such an expansion could bridge the gap between pure game-theoretic strategy and the rich, situated reality of human \mbox{decision-making}.

\section{Conclusions}

In this work, we introduced Mxplainer, a~novel framework that bridges the gap between the high performance of black-box agents and the critical need for interpretability in complex domains like Mahjong. Unlike existing imitation learning or post hoc explanation methods that often trade accuracy for transparency or struggle with high-dimensional state spaces, Mxplainer's unique approach constructs a parameterized, search-based agent whose components are directly convertible into an equivalent, trainable neural network. This core design allows it to accurately approximate sophisticated agents while retaining a human-understandable deduction process and strategically meaningful~parameters.

Our experiments demonstrate that Mxplainer successfully extracts and quantifies agent characteristics, such as fan and tile preferences, and~provides local explanations for individual decisions. The~framework's ability to manipulate agent behavior by tuning these parameters further validates that the learned values capture genuine strategic~inclinations.

Nevertheless, Mxplainer has limitations. Its approximation power is inherently constrained by the expressiveness of the hand-crafted search framework, and~the current use of uniform distributions for tile-drawing and discarding probabilities is a simplifying assumption that could be refined. Future work will focus on enhancing the model's expressiveness without sacrificing interpretability, exploring more sophisticated probability models, and, most importantly, generalizing the explanation template to a wider range of applications beyond Mahjong. We believe that the paradigm of converting a domain-specific, parameterized classical agent into a trainable network offers a promising path forward for building performant, transparent, and~trustworthy AI~systems.

\vspace{6pt}

\supplementary{The following supporting information can be downloaded at  \linksupplementary{s1}. Table S1: Correlation Between Learned Parameters and Tile Preference; Table S2: Correlation Between Learned Parameters and Fan Preference; Table S3: Interpretation of Deduction Process For Sampled Game State.}

\authorcontributions{Conceptualization, Lingfeng Li; methodology, Lingfeng Li; software, Lingfeng Li; validation, Lingfeng Li; investigation, Lingfeng Li, Yunlong Lu, Yongyi Wang; data curation, Lingfeng Li and Yunlong Lu; writing---original draft preparation, Lingfeng Li; writing---review and editing, Lingfeng Li, Yunlong Lu, Yongyi Wang, and Wenxin Li; visualization, Lingfeng Li; supervision, Wenxin Li; project administration, Wenxin Li. All authors have read and agreed to the published version of the manuscript.}

\funding{This research received no external funding. }

\dataavailability{Our data and codes, which encompass toy examples and comparative studies, are accessible at} \url{https://github.com/Lingfeng158/Mxplainer}.

\conflictsofinterest{The authors declare no conflicts of interest.  }

\appendixtitles{yes} 
\appendixstart
\appendix
\section[\appendixname~\thesection]{Mounter Car and Black Jack}\label{secA}
\begin{figure}[H]
\includegraphics[width=0.95\columnwidth]{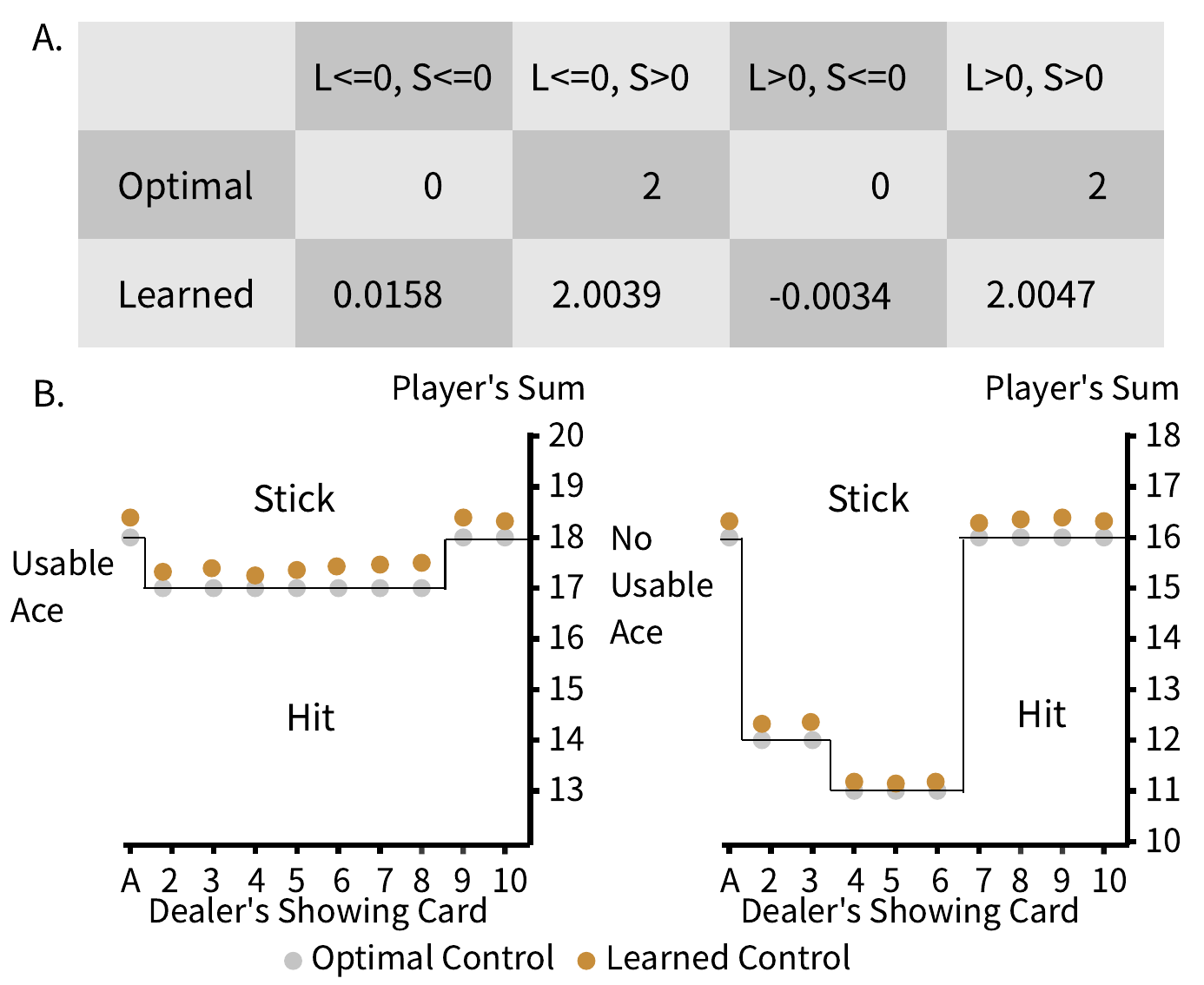} 
\caption{{Optimal} 
 parameters and learned parameters for (\textbf{A}) mountain car and (\textbf{B}) black~jack.}
\label{figdemo}
\end{figure}

In Mountain Car, the~observation space is location \textbf{L} and speed \textbf{S}, both of which can be positive or negative. The~action spaces are accelerations to the left or right. We can design a heuristic template $H$ that takes the sign of \textbf{L} and \textbf{S} as input, and~the heuristics, $\Theta$, are actions for these four states. The~transformation $T$ is also simple: We map \textbf{L} and \textbf{S} to the four states during data processing and output a one-hot mask for $\Theta$. When imitating the optimal strategy, the~learned $\Theta$ can be found in Figure~\ref{figdemo}A, which also achieves optimal~control.

In Blackjack, the~observation space includes usable ace \textbf{A}, dealer's first card \textbf{C}, and~player's sum of points \textbf{S}. The~action space is to hit or stick under each state. The~Heuristic Template $H$ is designed to act based on heuristic boundaries $\Theta$ of \textbf{S} for Hit or Stick for each (\textbf{A}, \textbf{C}) pair. The~transformation $T$ is again mapping (\textbf{A}, \textbf{C}) pairs to their masks. The~optimal strategy and the learned $\Theta$ is shown in Figure~\ref{figdemo}B, showing that the framework learns exactly the same behavior with the target model, i.e.,~the optimal~strategy.

\vspace{-6pt}

\unskip

\section[\appendixname~\thesection]{Rules and Examples of Transformation of Non-Differential Structures}\label{secB}

Boolean-to-Arithmetic Masking ($M$) is defined as a one-hot vector of dim $|n|$ for a conditional statement with $n$ branches. If~a conditional branch can be determined without parameters, $M$ can be computed and stored as input data; otherwise, $M$ can be computed at runtime. Computations for results $R$ are the same for all branches, but~only one result is activated through $R \cdot M$. 
         
The \textit{Padding Value} ($P$) is defined as the identity element of its related operator. Identity elements ensure that any computation paths with $P$ produce no-op results, and~they are required for the treatment of loops. For~example, if~the results of a loop are gathered through a summation operation, then the computation path with $P$ produces 0. If~the results are gathered through multiplication, then the path with $P$ produces 1. This is because for classical agents to be optimized and parallelized, loops must be set with upper limits, similarly to the maximum sequence length when training a recurrent neural network. $P$ needs to fill the iterations where there is no actual data. To~keep the Framework $F$ exactly the same as the transformed network $N$, all upper limits of loops need to be reflected in the original loops of $F$. Table~\ref{use_case} summarizes $M$ and $P$.

\begin{table}[H]

\caption{Examples of the transformation for conditional statements and loops. Note that classical representations of loops need to be modified with the same upper limit as the neural representations. $V^{\top[1,2]}$ is the example input with axis 1 and 2 transposed for efficient representation due to space~{constraints.} 
}
\label{use_case}
\centering
\small
\setlength{\tabcolsep}{2.7mm}
\begin{tabular}{@{}m{2.1cm}@{}m{5.2cm}@{}m{6.25cm}}
\toprule
Conditional & Classical Representation &  Neural Representation (Pytorch)\\ 
\midrule
&
\begin{lstlisting}[language=Python]
def foo(a,b):
    if P:
        r = a*3+b*4
        return r
    elif Q:
        return a*b
    else
        return a/b
    
\end{lstlisting}
&
\begin{lstlisting}[language=Python]
def forward(A, B, M):
    p = A*3 + B*4
    q = A*B
    o = A/B
    out = torch.cat([p,q,o], dim=1)
    return torch.sum(out*M, dim=1)
    
\end{lstlisting}

\\[-1.25em]

\midrule
Loop A& Classical Representation  & Neural Representation (Pytorch)\\ 
\midrule
\small\vspace{-10pt}
\makecell{
Note:\\
Results are\\
independent \\
between\\
Iterations.
}\vspace{-3pt}
&
\begin{lstlisting}[language=Python]
def bar1(A):
    res = []
    e = enumerate
    for i,a in e(A):
        if i>64:
            break
        r = 2*(a+1)
        res.append(r)
    p = np.prod
    return p(res) 
\end{lstlisting}
&
\begin{lstlisting}[language=Python]
def forward(V):
    V1 = V.view(n*64, 2)
    O = V1[:,0]*2*(V1[:,1]+1)
    # convert padding values to 1
    # for the final torch.prod()
    O += 1-V1[:, 0]
    O1 = O.view(n, 64)
    return torch.prod(O1, dim=1)
    
\end{lstlisting}
\\[-1.25em]
\midrule
Loop B& Classical Representation& Neural Representation (Pytorch) \\
\midrule
\small\vspace{-22pt}
\makecell{
Note:\\
Results are\\
dependent \\
between\\
Iterations.
}&

\begin{lstlisting}[language=Python]
def bar2(A):
    res = []
    e = enumerate
    for i,a in e(A):
        if i>64:
            break
        r=a
        if i!=0:
            r += res[-1]
        res.append(r)
    return sum(res) 
\end{lstlisting}
&
\begin{lstlisting}[language=Python]
def forward(V):
    res = []
    for i in range(64):
        if i==0:
            r = V[:, i, 1]
        else:
            # computing real data
            r = res[-1]+V[:, i, 1]
        # set zero for padding values
        r*=V[:, i, 0]
        res.append(r)
    res = torch.stack(res)
    res = torch.sum(res, dim=0)
    return res
    
\end{lstlisting}
\\[-1.25em]
\bottomrule
\end{tabular}

\label{tab:transformation_example}

\end{table}

We provide a lemma to show that $M$ and $P$ result in equivalent $F$ and $N$. 

\begin{Lemma}{Semantic Equivalence under Identity Padding and Masking}

Let $F$ be a classical computation framework containing loops with a fixed upper limit $L$ and conditional branches. Let $N$ be its neural network transformation, implemented~using the following: 
\begin{itemize}
    \item A padding value $P$ that is the identity element for the accumulation operation in a loop (e.g., 0 for summation, 1 for product). 
    \item A Boolean mask $M$ that is a one-hot vector for conditional branches, selecting the active computational path. 
\end{itemize}
{If} 
 the following conditions~hold, {then} for~any input S, the~output of $N(S)$ is identical to the output of $F(S)$:
\begin{itemize}
    \item All loops in $F$ are padded to the same upper limit $L$ as in $N$, with~padding elements set to $P$. 
    \item All branch conditions in $F$ are deterministically mapped to their corresponding one-hot masks $M$ in $N$. 
    \item The accumulation operations (e.g., sum, prod) in $F$ are associative and commutative, and~$P$ is their true identity element. 
\end{itemize}

\end{Lemma}
\begin{proof}
\paragraph{Base Case} For elementary operations (e.g., arithmetic operations on tensors that do not involve control flow), the~transformation 
$T$ is the identity function. By~definition, $N(S)=F(S)$ for these operations.

\begin{enumerate}[leftmargin=0em,labelsep=4mm]
\item[] {{Conditional} 
 Branching} 
\end{enumerate}

Consider a conditional statement in $F$ with $n$ branches.
\begin{listing}[H]
\caption{Branching in Python}%
\label{lst:proof_step1_F}%
\begin{lstlisting}[language=Python]
# Classical Representation F
if condition_1:
    result = f_1(S)
if condition_2:
    result = f_2(S)
...
if condition_n:
    result = f_n(S)
    
\end{lstlisting}
\end{listing}

This is transformed in $N$ into the following.

\begin{listing}[H]
\caption{Branching in Neural Network}%
\label{lst:proof_step1_N}%
\begin{lstlisting}[language=Python]
# Neural Representation N
M = one_hot(condition_1, condition_2) #Mask Vector
R = stack([f_1(S), f_2(S), ..., f_n(S)]) # All Branch Results
result = sum(R * M)  # Mask application    
\end{lstlisting}
\end{listing}
By Condition 2, the~branch conditions are deterministically mapped to a one-hot mask $M$, where exactly one element is $M[i]=1$, and all others are 0. The~stack operation computes all branch results $f_1(S), f_2(S), ...,f_n(S)$ in parallel. The~operation $sum(R * M)$ performs an element-wise multiplication of the result tensor $R$ with the mask $M$. Since $M$ is one-hot, this operation selects the $i$-th branch result $f_i(S)$ and sums it with zeros from all other branches. Therefore, the result in $N$ is identical to the result in $F$, as~both equal $f_i(S)$.

\paragraph{Fixed-Bound Loops}
Consider a loop in $F$ with an upper bound $L$, which accumulates results using an operator $\oplus$ with identity element $P$:

\begin{listing}[h][mathescape=true]
\caption{Fixed Bound Loops in Python}%
\label{lst:proof_step2_F}%
\begin{lstlisting}[language=Python, mathescape=true]
# Classical Representation F
result = P                             # Initialize with identity
for i in range(L):
    if i < actual_length:
        element = get_element(i, S)
        result = result $\oplus$ element
    # Implicit else: pad with identity~P
    
\end{lstlisting}
\end{listing}

This is transformed in $N$ into the following.
\begin{listing}[h!]
\caption{Fixed Bound Loops in Python}%
\label{lst:proof_step2_N}%
\begin{lstlisting}[language=Python, mathescape=true]
# Neural Representation N
all_elements = get_all_elements(S, L)  # Padded with P to length L
result = reduce(all_elements, $\oplus$)       # Parallel reduction with $\oplus$
    
\end{lstlisting}
\end{listing}

By Condition 1, the~loop in $F$ is explicitly padded to length $L$ with the identity element $P$. The~neural version $N$ directly operates on this padded sequence \textit{all\_elements}. By~Condition 3, the~accumulation operation $\oplus$ is associative and commutative. The~fundamental property of an identity element $P$ is that for any $x$, $x \oplus P = P \oplus x = x$. Therefore, performing the reduction $\oplus$ over the padded sequence \textit{all\_elements} is equivalent to performing the sequential loop in $F$: The contributions of the real elements are unchanged, and~the padding elements $P$ leave the accumulation result invariant. Thus, the~result computed by $N$ is identical to that computed by $F$.
\end{proof}

This lemma provides the formal basis for our transformation. The~padding value $P$ ensures that padded elements do not alter the result of the accumulation, while the mask M ensures that only the correct branch result is~propagated. 

To empirically verify the equivalence between framework $F$ and network $N$ under varied conditions, we conducted large-scale differential testing. We sampled 10,000 game states from our database and generated random parameter weights for $N$, which were then directly transferred to $F$. The~outputs of both systems—specifically, the~values computed by Algorithm 1/Listing 1 and the tile preferences computed by Algorithm 2/Listing 2—were compared for every state. The~results were identical across all 10,000 test cases, confirming the semantic equivalence of $F$ and $N$. \\

\begin{adjustwidth}{-\extralength}{0cm}

\reftitle{References}

\PublishersNote{}
\end{adjustwidth}
\end{document}